\documentclass[10pt,twocolumn,letterpaper]{article}

\usepackage{cvpr}      

\usepackage{cuted} 

\usepackage{currfile} 

\usepackage{caption} 
\definecolor{cvprblue}{rgb}{0.21,0.49,0.74}
\usepackage[pagebackref,breaklinks,colorlinks,allcolors=cvprblue]{hyperref}

\def\confName{CVPR}
\def\confYear{2026}

\title{PLaCA: Learning Patch-Level Future Latent Change for Action Generation}

\author{
Zhan Yi \qquad
Yuchen Wang \qquad
Chao Li \qquad
Lei Hong \qquad
Boyuan Yang$^{*}$\\
School of Robotics and Automation, Nanjing University\\
Suzhou, Jiangsu, China\\
{\normalfont\mdseries\small\ttfamily
njuyizhan@smail.nju.edu.cn,
wyc\_nju@smail.nju.edu.cn,
lichao\_nju@smail.nju.edu.cn,}\\
{\normalfont\mdseries\small\ttfamily
leihong@smail.nju.edu.cn,
yby@nju.edu.cn
}\\
{\normalfont\small $^{*}$Corresponding author.}
}

\begin{document}
\maketitle

\begin{abstract}
Vision-Language-Action (VLA) models generate robot
actions from visual observations and language
instructions, but effective manipulation also
requires anticipating how the environment will
change. Existing world-action models explore future
image prediction and latent dynamics modeling.
However, generating complete future images can be
computationally expensive, while highly compressed
latent representations may obscure local changes.
To address these challenges, we propose PLaCA,
a world-to-action framework built around future
latent change. PLaCA directly predicts spatially
structured patch-level future latent change from
the current multimodal input, preserving local
information about expected environmental changes.
We introduce patch-level change-aware supervision
to emphasize regions undergoing larger changes
and a future-to-action interface that explicitly
incorporates predicted future latent change into
action generation.
Experiments on LIBERO-Long show that PLaCA achieves
a 79.0\% success rate, improving upon the Action-only
baseline by 5.0 percentage points. On CALVIN, PLaCA
improves the average sequence length from 2.784 to
3.564. These improvements are achieved while updating
only 2.95\% of the model parameters during the second
training stage, demonstrating the effectiveness of
future latent change modeling for downstream robotic
manipulation.
\end{abstract}

\section{Introduction}

Vision-Language-Action (VLA) models typically generate robot
actions from current visual observations and language
instructions~\cite{rt12022,octo2024,rdt2025}. However, in manipulation tasks such as grasping,
pushing, and placing, effective action generation requires
more than understanding the current environment. A policy
must also consider how its actions may change object positions,
contact relations, and local environmental states. To
incorporate such future information, recent studies have
explored integrating world modeling with VLA policies,
motivating the development of world-action modeling (WAM)
~\cite{cen2025worldvla,WorldAgen}.

Existing approaches represent future information in different
ways. Some predict future images or videos~\cite{genie2024,robodreamer2024},
while others model future states or temporal dynamics in latent
space~\cite{dreamerv32025,vjepa22025}.
Recent studies have further explored compact latent motion
representations~\cite{StaMo,CoWVLA,yang2026como}.
Latent-space modeling has become a widely adopted approach
in world modeling and robot learning~\cite{dreamerv32025,vjepa22025,CoWVLA}.
While pixel-level future prediction provides rich visual
details, generating complete future images often involves
reconstructing substantial redundant visual content,
leading to additional computational costs~\cite{ivideogpt2024,CoWVLA}.
In contrast, latent-space modeling avoids explicit pixel-level
reconstruction, yet choosing an appropriate future representation
for robotic manipulation remains an important question.
For action generation, it may not be necessary to recover
the complete future state or all intermediate dynamics.
Instead, a policy may benefit from information about
how the environment is expected to change.

\begin{figure}[t]
    \centering
    \includegraphics[width=\linewidth]{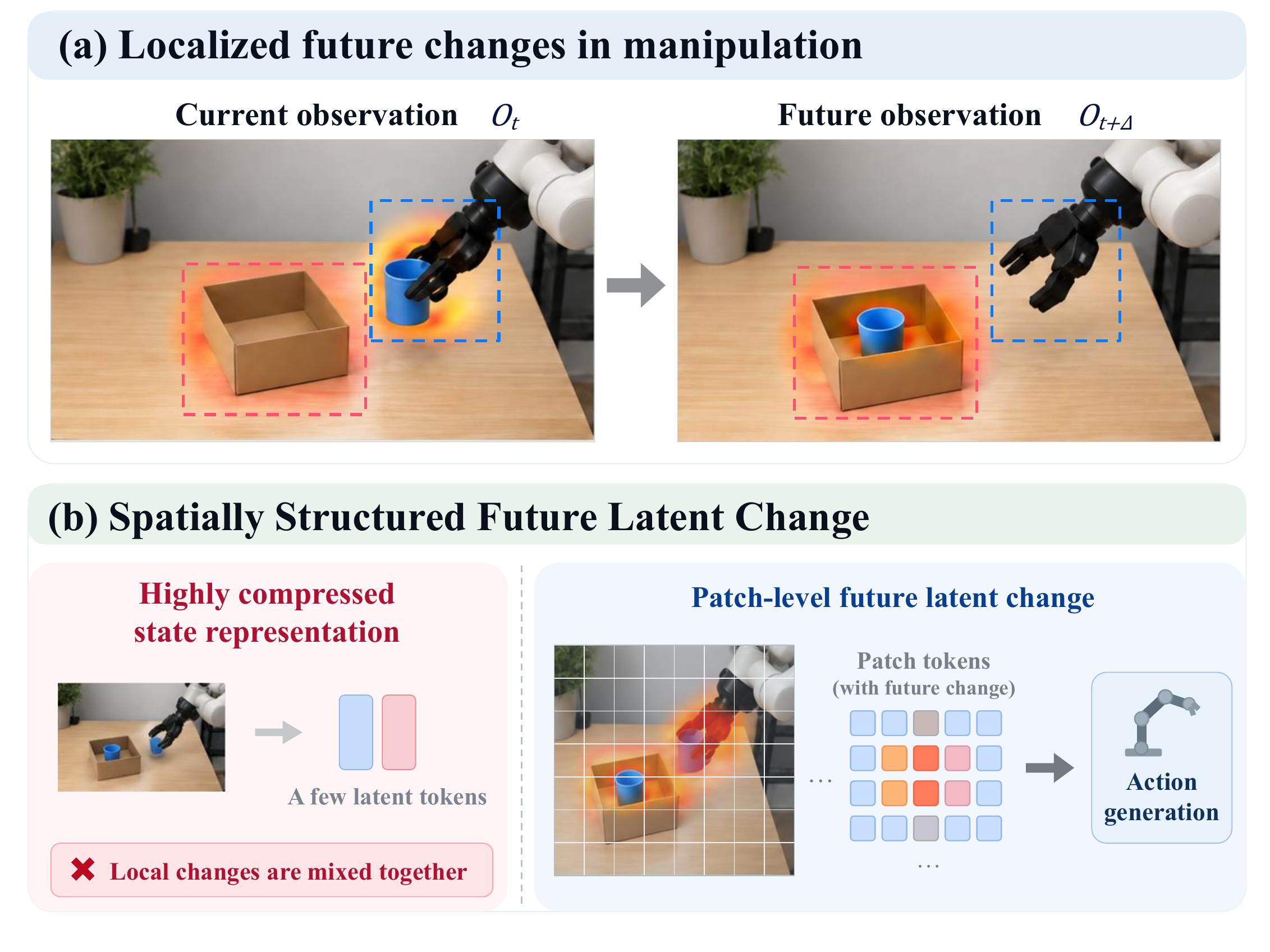}
    \caption{
    \textbf{Motivation for spatially structured future
    latent change.}
    (a) In robotic manipulation, future changes are often
    concentrated around the robot arm and manipulated
    objects, while much of the background remains unchanged.
    (b) Highly compressed state representations do not
    explicitly preserve the spatial locations of these
    changes, whereas patch-level future latent change
    retains their spatial structure and provides future
    information for action generation.
    }
    \label{fig:intro}
\end{figure}

Recent work has shown that differences between latent states
can capture information related to robot motion. StaMo~\cite{StaMo} learns
highly compressed latent state representations and uses
the latent displacement between states to represent robot
motion. It also explores latent motion prediction for world
modeling. However, latent representations differ
in how they preserve spatial information. As illustrated in
Fig.~\ref{fig:intro}(a), when a robot places an object into
a box, the changes are mainly concentrated around the robot
arm and the object, while much of the background remains
unchanged. Highly compressed state representations can encode
such changes, but they do not explicitly preserve the
correspondence between spatial locations and their latent
changes. In contrast, retaining the patch-level spatial
structure allows future changes at different image locations
to be represented separately, as shown in
Fig.~\ref{fig:intro}(b). Moreover, the magnitude of future
change can vary considerably across patches, while uniform
patch supervision does not specifically emphasize regions
undergoing larger changes. This motivates our first research
question: \emph{How can we preserve the spatial structure
of future latent change and learn local changes more
effectively by accounting for their different magnitudes?}

Another important question is how future modeling can be effectively coupled with action generation. Prior work has explored different approaches, including jointly learning future prediction and action policies within shared architectures~\cite{WorldAgen,CoWVLA}, using predicted future images or video plans to guide downstream control~\cite{cotvla2025,videolanguageplanning2024}, and leveraging learned latent or interaction dynamics for action inference~\cite{VideoWorld,DynBridge}. These studies provide different ways of using future information to inform robotic behavior. For the spatially structured future latent change considered in this work, the predicted representation preserves local information about expected environmental changes across image regions. This motivates investigating how such spatially organized predictions can be incorporated into action representations to support action generation. We therefore pose our second research question: \emph{How can predicted future latent change be effectively used to support action generation?}

To address these questions, we propose \textbf{PLaCA},
a world-to-action framework built around
\emph{future latent change}. PLaCA directly predicts
spatially structured patch-level future latent change
from the current multimodal input and employs
\emph{patch-level change-aware supervision} to emphasize
regions with larger changes. We further introduce a
\emph{future-to-action interface} that incorporates the
predicted future latent change into the action representation,
allowing it to directly contribute to full action-chunk
generation.

We evaluate PLaCA on LIBERO-Long, achieving a 79.0\% task success rate. Ablation studies demonstrate that preserving fine-grained spatial structure, employing change-aware supervision, and incorporating predicted future latent change into action generation through the future-to-action interface contribute to improved downstream manipulation success rates.

Our main contributions are summarized as follows:
\begin{itemize}
    \item We investigate \emph{patch-level future latent change
    modeling}, which directly predicts future latent change
    from the current multimodal input while preserving its
    spatial structure, providing local future information
    for action generation.

    \item We propose \emph{patch-level change-aware supervision},
    which adjusts the supervision weights according to the
    magnitude of ground-truth future latent change.
    Ablation results show that this approach improves
    closed-loop manipulation performance compared with
    uniform patch supervision.

    \item We design an explicit \emph{future-to-action interface}
    that incorporates the final predicted future latent
    change into the action representation, allowing future
    information to directly contribute to full action-chunk
    generation.
\end{itemize}

\section{Related Work}

\subsection{World-Action Modeling}

Vision-Language-Action (VLA) models unify visual observations, language
instructions, and robot actions within end-to-end policies, extending
large-scale vision-language models toward general-purpose robotic
control~\cite{VLA,kim2024openvla,black2025pi0,rt12022,openx2024,octo2024,rdt2025}. While conventional VLAs
primarily focus on predicting continuous actions or action sequences from
current multimodal observations~\cite{pi052025,gr00tn12025,smolvla2025,tinyvla2024,vlaadapter2026}, recent studies have increasingly incorporated
future modeling, enabling policies to reason not only about \emph{what to do}
but also about \emph{what may happen} after an action is
executed~\cite{cen2025worldvla,zhu2025uwm,cotvla2025,videolanguageplanning2024}. This trend has gradually extended
VLA research from direct observation-to-action mapping toward joint modeling of
environmental dynamics and robot behavior.

One line of work models future environmental changes through
image or video prediction~\cite{ivideogpt2024}, while other
approaches jointly learn future modeling and action
generation~\cite{cen2025worldvla,zhu2025uwm}.
WorldAgen performs future-state prediction and action prediction
within a unified Transformer, allowing world modeling and policy
learning to share internal representations~\cite{WorldAgen}.
CoT-VLA uses predicted future images as intermediate visual guidance
for subsequent action generation~\cite{cotvla2025}. DynBridge learns interaction dynamics that connect future
interaction modeling with robot control~\cite{DynBridge}, while Motus further
integrates latent action modeling, world modeling, and action generation into a
unified framework, using shared latent representations to characterize the
relationship between robot behavior and environmental changes~\cite{Motus}.
DreamVLA combines dynamic-region-guided future information with spatial and
semantic cues for action generation~\cite{zhang2025dreamvla}, while Mantis
learns latent visual foresight with a separate prediction
head~\cite{yang2026mantis}. Collectively, these studies reflect a growing trend
in which future modeling is increasingly coupled with action generation through
shared or explicitly connected representations. We focus on directly predicting
spatially structured future latent change and incorporating its predicted tokens
into the action representation through gated cross-attention.

\subsection{Latent Dynamics}

Future image and video prediction provides a detailed
description of environmental changes, but generating full
future frames can be computationally expensive~\cite{robodreamer2024,genie2024}. Recent
studies have therefore explored latent state, latent motion,
and latent action representations for future modeling and
robot control~\cite{bu2025univla}~\cite{lapa2024,vjepa2024,vjepa22025,dreamerv32025}.

One line of work focuses on representing dynamics in
latent space. CoWVLA separates structural and motion
information and learns a continuous latent motion chain
to represent temporal evolution~\cite{CoWVLA}. CoMo learns
continuous latent motion by combining temporal differences
with contrastive learning~\cite{yang2026como}. StaMo takes
a different approach by learning highly compressed latent
state representations from individual observations. It
shows that differences between these states contain robot
motion information and can support action decoding.
StaMo further uses state and motion representations as
prediction targets for VLA world modeling and explores
latent-motion co-training~\cite{StaMo}. These studies
explore different ways to capture temporal changes in
latent space, including continuous motion representations
and differences between compact latent states.

Another line of work investigates how latent dynamics
can support action generation. VideoWorld learns discrete
latent codes that represent future visual changes and
uses the predicted codes together with visual information
for inverse dynamics-based action decoding~\cite{VideoWorld}.
Latent Bridge instead predicts continuous feature changes
to update latent representations, allowing the policy to
reduce repeated VLM computation during action
generation~\cite{LatentBridge}. These approaches illustrate
different ways of using learned latent dynamics to
support action generation.

Our work focuses on learning spatially structured
future latent change and using the predicted
representation for action generation.

\section{Method}
\label{sec:method}

\subsection{Overview}
\label{sec:method_overview}

\begin{figure*}[t]
    \centering
    \includegraphics[width=\textwidth]{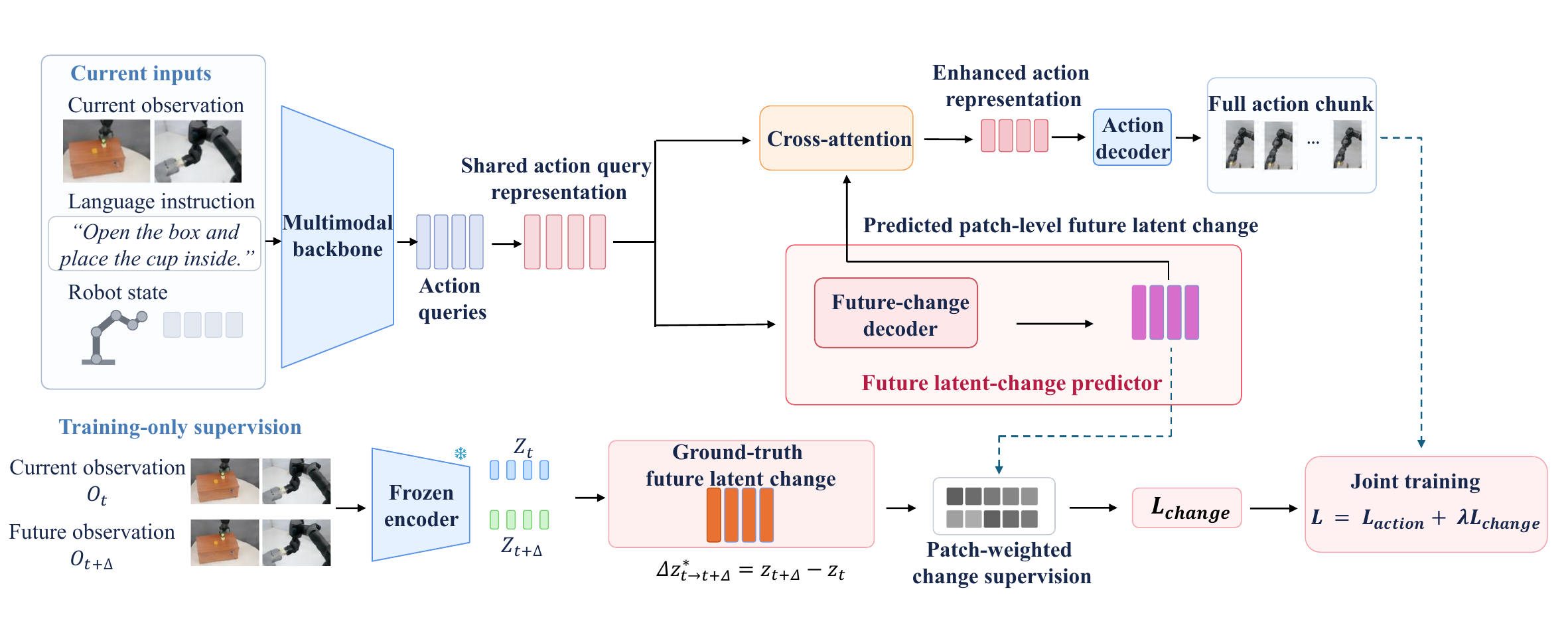}
    \caption{
    \textbf{Overview of our world-to-action framework.}
    The shared multimodal backbone produces action-query
    representations, from which a future-change decoder
    predicts patch-level future latent change.
    During training, a frozen visual encoder provides
    change-aware supervision using current and future
    observations. The predicted future latent change
    is then incorporated into the action representation
    through a gated future-to-action interface to
    generate the full action chunk. Future observations
    are not required during inference.
    }
    \label{fig:method}
\end{figure*}

We consider robotic manipulation conditioned on multimodal
context. At time step $t$, the current observation is
$O_t=\{O_t^c\}_{c=1}^{C}$, where $C$ is the number of camera
views. The language instruction, robot state, and available
observation-action history are denoted by $\ell$, $s_t$,
and $\mathcal{H}_t$, respectively. Together, they form
the multimodal input
$\mathcal{X}_t=(O_t,\ell,s_t,\mathcal{H}_t)$.
Our goal is to predict a full action chunk
$A_t=\{a_t,\ldots,a_{t+H-1}\}\in\mathbb{R}^{H\times d_a}$,
where $H$ is the action horizon and $d_a$ is the action
dimension~\cite{act2023,diffusionpolicy2023,openvlaoft2025}. We use $H=3$ and set the future prediction
horizon to $\Delta=3$, aligning the future observation
with the state reached after executing the action chunk.

As shown in Fig.~\ref{fig:method}, our framework consists
of a shared multimodal backbone, a future-change decoder,
and a future-to-action interface. The shared backbone
$\mathcal{B}_{\theta}$ processes the multimodal input
together with action-query tokens $Q^a$ to obtain an
initial action representation $H_t^{(0)}$~\cite{attention2017,qwen22024}.
A residual adapter produces the shared action-query
representation~\cite{vlaadapter2026,lora2022}:
\begin{equation}
\begin{aligned}
H_t^{(0)} &= \mathcal{B}_{\theta}
    (\mathcal{X}_t,Q^a),\\
H_t^a &= H_t^{(0)}
    +\alpha\,\operatorname{Adapter}(H_t^{(0)}),
\end{aligned}
\label{eq:shared_action_rep}
\end{equation}
where $\alpha$ is a zero-initialized learnable scalar.
The same representation $H_t^a$ is used for future
latent change prediction and action generation.

The overall prediction process is formulated as
\begin{equation}
\begin{aligned}
\widehat{\Delta Z}_t
    &= \mathcal{D}_f(Q^f,H_t^a),\\
\widetilde{H}_t^a
    &= \mathcal{F}_{\phi}
       (H_t^a,\widehat{\Delta Z}_t),\\
\widehat{A}_t
    &= \mathcal{D}_a(\widetilde{H}_t^a),
\end{aligned}
\label{eq:overall_framework}
\end{equation}
where $\mathcal{D}_f$ denotes the future-change decoder,
$Q^f$ denotes learnable change queries,
$\mathcal{F}_{\phi}$ is the future-to-action interface,
and $\mathcal{D}_a$ is the action decoder.
The predicted future latent change
$\widehat{\Delta Z}_t$ provides both a future prediction
target during training and conditioning information
for action generation.

\subsection{Patch-Level Future Latent Change Modeling}
\label{sec:future_latent_change}

We model future information as changes between the
current and future latent states while preserving
their patch-level spatial structure. Instead of
representing the change using a single global feature,
we predict a separate latent change vector for each
patch location in every camera view.

\noindent\textbf{Ground-Truth Future Latent Change.}
As illustrated in Fig.~\ref{fig:patch_change_supervision}(a),
we construct the ground-truth future latent change
from the current observation $O_t$ and future observation
$O_{t+\Delta}$. A frozen visual encoder $E_{\mathrm{vis}}$
extracts patch-level latent state representations
~\cite{vit2021,mae2022}. For the $p$-th patch of
camera view $c$, we define :

\begin{equation}
\begin{aligned}
z_t^{c,p}
    &= E_{\mathrm{vis}}(O_t^c)_p,\\
z_{t+\Delta}^{c,p}
    &= E_{\mathrm{vis}}(O_{t+\Delta}^c)_p,\\
\Delta z_t^{*,c,p}
    &= z_{t+\Delta}^{c,p}-z_t^{c,p}.
\end{aligned}
\label{eq:future_change_target}
\end{equation}

Collecting the differences across all camera views
and patch locations yields
$\Delta Z_t^*\in\mathbb{R}^{C\times P\times d_z}$,
where $P=H_pW_p$ denotes the number of patches
per view, $H_p$ and $W_p$ are the height and
width of the patch grid, and $d_z$ is the
latent feature dimension.

We use a frozen MAE ViT encoder~\cite{mae2022,vit2021} to construct these
targets. Unlike the compact visual tokens supplied
to the policy backbone, the supervision targets retain
the complete patch grid without spatial pooling.
Each difference is computed between features at the
same image patch location, preserving the spatial
organization of the latent representation.

\noindent\textbf{Future Latent Change Prediction.}
To predict this representation, we introduce
$Q^f\in\mathbb{R}^{(CP)\times d}$ as learnable
change queries, with each query assigned to a
camera view and patch location. Here, $d$ denotes
the hidden dimension of the future-change decoder. A Transformer-based
future-change decoder uses these queries~\cite{attention2017,perceiverio2022} to attend
to the shared action-query representation $H_t^a$,
aggregating information from the current multimodal
context.

The decoder outputs are projected into the visual
latent space to obtain
\begin{equation}
\widehat{\Delta Z}_t
    =\mathcal{D}_f(Q^f,H_t^a)
    \in\mathbb{R}^{C\times P\times d_z}.
\label{eq:future_change_prediction}
\end{equation}
The predicted vectors retain their camera and patch
indices and are supervised at the corresponding
spatial locations. Importantly, the decoder directly
predicts the continuous future latent change rather
than first reconstructing a future latent state.
This prediction is subsequently used by the
future-to-action interface.

\subsection{Patch-Level Change-Aware Supervision}
\label{sec:change_supervision}

\begin{figure*}[t]
    \centering
    \includegraphics[width=\textwidth]{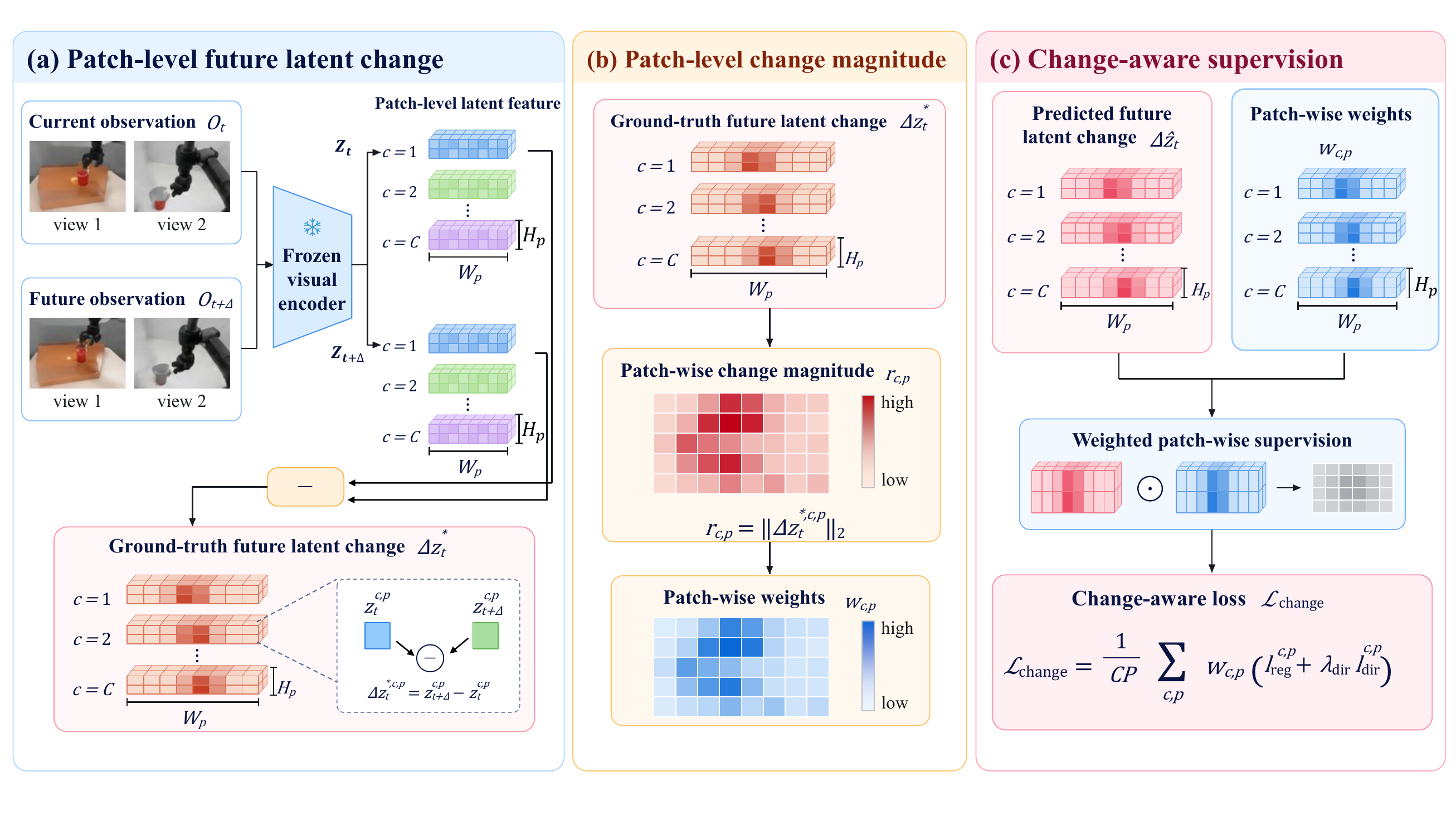}
    \caption{
    \textbf{Patch-level future latent change and change-aware supervision.}
    (a) A frozen visual encoder extracts patch-level features from
    current and future observations. Their differences at
    corresponding patch locations define the ground-truth
    future latent change $\Delta Z_t^*$, preserving the spatial
    structure across camera views.
    (b) The magnitude of each ground-truth change vector
    is used to compute patch-wise supervision weights,
    assigning larger relative weights to patches with
    greater changes.
    (c) The predicted future latent change
    $\widehat{\Delta Z}_t$ is supervised against
    $\Delta Z_t^*$ using weighted regression and
    directional losses. All patches contribute to
    the training objective with different weights.
    }
    \label{fig:patch_change_supervision}
\end{figure*}

Future latent change is generally distributed
unevenly across image regions. While some patches
undergo substantial changes, others remain relatively
stable over the prediction horizon. To account for
these differences, we introduce patch-level
change-aware supervision that assigns larger relative
weights to patches with greater ground-truth changes. Figure~\ref{fig:patch_change_supervision}(b)
illustrates how patch-wise change magnitudes
are used to determine the supervision weights.

For each patch, we measure its change magnitude as
$r_{c,p}=\|\Delta z_t^{*,c,p}\|_2$ and compute the
mean magnitude within each sample:
\begin{equation}
\bar r=\frac{1}{CP}\sum_{c,p}r_{c,p}.
\label{eq:mean_change_magnitude}
\end{equation}
We then construct normalized supervision weights:
\begin{equation}
\begin{aligned}
\widetilde w_{c,p}
&=\operatorname{clip}\left(
1+\eta\left(
\frac{r_{c,p}}{\max(\bar r,\epsilon)}-1
\right),
w_{\min},w_{\max}
\right),\\
w_{c,p}
&=\frac{\widetilde w_{c,p}}
{\max\left(
\frac{1}{CP}\sum_{c',p'}\widetilde w_{c',p'},
\epsilon
\right)}.
\end{aligned}
\label{eq:patch_weights}
\end{equation}

Here, $\eta$ controls the strength of change-aware
weighting. We use $\eta=0.75$, $w_{\min}=0.25$,
and $w_{\max}=4$, with $\epsilon=10^{-8}$ for
numerical stability. The weights are normalized
after clipping to maintain a mean of one.
All patches remain supervised, including those
with small changes.

We combine a SmoothL1 regression loss with a
cosine directional loss to supervise the predicted
future latent change. For each patch, these are
defined as
\begin{equation}
\begin{aligned}
\ell_{\mathrm{reg}}^{c,p}
&=\frac{1}{d_z}\sum_{j=1}^{d_z}
\operatorname{SmoothL1}
\left(
\widehat{\Delta z}_{t,j}^{c,p},
\Delta z_{t,j}^{*,c,p}
\right),\\
\ell_{\mathrm{dir}}^{c,p}
&=1-\operatorname{cos}
\left(
\widehat{\Delta z}_t^{c,p},
\Delta z_t^{*,c,p}
\right),
\end{aligned}
\label{eq:patch_change_losses}
\end{equation}
where cosine similarity is computed with
numerically stabilized vector norms.
The weighted future latent change objective is
\begin{equation}
\mathcal{L}_{\mathrm{change}}
=\frac{1}{CP}\sum_{c,p}w_{c,p}
\left(
\ell_{\mathrm{reg}}^{c,p}
+\lambda_{\mathrm{dir}}\ell_{\mathrm{dir}}^{c,p}
\right),
\label{eq:future_change_loss}
\end{equation}

where $\lambda_{\mathrm{dir}}=0.2$.
As illustrated in
Fig.~\ref{fig:patch_change_supervision}(c),
the patch-wise weights adjust the contributions
of the regression and directional losses.
These weights are used only during training
and are not required during inference.

\subsection{Future-to-Action Interface}
\label{sec:future_to_action}

Predicting future latent change provides information
about the expected environmental changes. To make
this information available to the action-generation
pathway, we introduce a future-to-action interface
that conditions the action representation on the
predicted patch-level future latent change.

\noindent\textbf{Future-Change Conditioning.}
We use the final predicted change vectors
$\widehat{\Delta Z}_t$, rather than the hidden
features of the future-change decoder, as the
input to the interface. After flattening the
camera and patch dimensions, a learnable
projection $\mathcal{P}_{\phi}$ maps the predicted
change into the action feature space:
\begin{equation}
F_t=\mathcal{P}_{\phi}
\left(\widehat{\Delta Z}_t\right)
\in\mathbb{R}^{(CP)\times d}.
\label{eq:future_projection}
\end{equation}

We then use cross-attention to incorporate the
predicted future latent change into the action
representation~\cite{attention2017,flamingo2022}. The shared action-query
representation $H_t^a$ serves as the Query,
while $F_t$ serves as the Key and Value:
\begin{equation}
U_t=\operatorname{LN}\left(
\operatorname{MHA}
\left(
Q=H_t^a,\,
K=F_t,\,
V=F_t
\right)
\right).
\label{eq:future_cross_attention}
\end{equation}
This allows each action query to aggregate
information from the predicted patch-level
future latent change.

\noindent\textbf{Gated Residual Integration.}
To control the contribution of predicted future
information, we employ a learnable scalar gate
$\beta$ and update the action representation
through residual fusion~\cite{flamingo2022}:
\begin{equation}
\begin{aligned}
\widetilde H_t^a &= H_t^a+\beta U_t,\\
\widehat A_t &=\mathcal{D}_a(\widetilde H_t^a).
\end{aligned}
\label{eq:gated_future_action}
\end{equation}
The gate $\beta$ is initialized to zero and
learned jointly with the other trainable modules.
It is independent of the residual-adapter
coefficient $\alpha$ in Eq.~\eqref{eq:shared_action_rep}.

The resulting action representation is decoded
into the full action chunk. Thus, the predicted
future latent change contributes directly to
action generation through a separate conditioning
pathway, rather than serving only as an
auxiliary prediction target.

\subsection{Joint Learning and Initialization}
\label{sec:joint_learning}

We adopt a two-stage training strategy.
First, we train an action-only policy using
behavioral cloning~\cite{act2023,diffusionpolicy2023,cdp2025} to establish an
observation-to-action mapping. We then initialize
the joint model from this policy and introduce
the future-change decoder and future-to-action
interface. During the second stage, the visual
and language encoders and the original backbone
weights remain frozen, while lightweight
backbone adapters, future-modeling modules,
the interface, and action-generation modules
are optimized.

\noindent\textbf{Action Generation Objective.}
We supervise the predicted full action chunk
using SmoothL1 loss for arm actions and
binary cross-entropy for gripper states:
\begin{equation}
\mathcal{L}_{\mathrm{action}}
=
\mathcal{L}_{\mathrm{arm}}^{\mathrm{SmoothL1}}
+\lambda_g
\mathcal{L}_{\mathrm{gripper}}^{\mathrm{BCE}},
\label{eq:action_loss}
\end{equation}
where $\lambda_g$ controls the contribution
of the gripper loss.

Training samples contain two consecutive action
chunks~\cite{act2023,fast2025} associated with observation endpoints
$O_0$, $O_3$, and $O_6$. Predicted future latent change from both chunks is incorporated into their respective action representations. Both chunks receive action supervision,
while direct future latent change supervision is applied
to the latest chunk, which corresponds to the action
chunk executed during closed-loop inference.
Specifically, we use
$\Delta Z_3^* = E_{\mathrm{vis}}(O_6) - E_{\mathrm{vis}}(O_3)$
as the prediction target. 

\noindent\textbf{Joint Optimization.}
The complete second-stage training objective is
\begin{equation}
\mathcal{L}_{\mathrm{total}}
=
\mathcal{L}_{\mathrm{action}}
+\lambda_{\mathrm{change}}
\mathcal{L}_{\mathrm{change}},
\label{eq:joint_loss}
\end{equation}
where $\lambda_{\mathrm{change}}$ controls the
contribution of future latent change supervision.

The future-change decoder receives direct
supervision from $\mathcal{L}_{\mathrm{change}}$.
Because its predictions are incorporated into
the action-generation pathway without detachment,
the action objective can also update the
future-change decoder as the gate $\beta$
learns a nonzero value. During inference,
the model predicts future latent change
and generates the full action chunk using
only the available multimodal context,
without access to future observations.

\section{Experiments}
\label{sec:experiments}

We evaluate PLaCA on two robotic manipulation benchmarks,
LIBERO-Long and CALVIN, and compare its performance with
representative methods. We further conduct ablation studies
to examine the contributions of patch-level spatial structure,
change-aware supervision, and the future-to-action interface.

\subsection{Experimental Setup}
\label{sec:experimental_setup}

\noindent\textbf{Benchmarks and Datasets.}
LIBERO-Long~\cite{libero,openvlaoft2025,spatialvla2025}
consists of 10 long-horizon manipulation tasks.
We use 50 expert demonstrations per task, totaling
500 trajectories, with 450 for training and 50 for
validation. All variants share the same trajectory-level
split. Each task is evaluated with 20 closed-loop rollouts,
yielding 200 episodes per model.

CALVIN~\cite{mees2022calvin} evaluates long-horizon language-conditioned robotic
manipulation across different environments. We adopt the
ABC$\rightarrow$D setting, training on environments A,
B, and C and evaluating on the unseen environment D.
We evaluate 1,000 task sequences, each containing five
consecutive subtasks, and report the Average Sequence
Length (ASL).

\noindent\textbf{Implementation Details.}
We build PLaCA on a Qwen-based Transformer backbone~\cite{qwen22024,attention2017}
with frozen MAE visual and CLIP language encoders~\cite{mae2022,clip2021}.
Following the two-stage training procedure described
in Sec.~\ref{sec:method}, all future-modeling variants
are initialized from the same Action-only checkpoint.
During the second stage, the pretrained backbone
weights remain frozen, with rank-8 LoRA~\cite{lora2022} applied to
the last six Transformer layers. The future-change
decoder, future-to-action interface, and action-generation
modules are jointly optimized.


\subsection{Evaluation on Simulation Benchmarks}
\label{sec:simulation_results}

\noindent\textbf{Results on LIBERO-Long.}
We evaluate PLaCA on the LIBERO-Long benchmark, which
consists of 10 long-horizon manipulation tasks involving
object grasping and placement, container manipulation,
and multi-stage task execution. We evaluate all 10 tasks
with 20 closed-loop rollouts per task, resulting in 200
episodes in total. We report the average task success rate.

\begin{table}[t]
    \centering
    \caption{
    \textbf{Comparison on LIBERO-Long.}
    Params denotes the reported model size.
    Results from prior methods are taken from their
    respective papers. 
    $\dagger$ PDF uses a 7B OpenVLA backbone with an
    additional 9M trainable parameters.
    }
    \label{tab:libero_long_comparison}
    \setlength{\tabcolsep}{9pt}
    \renewcommand{\arraystretch}{1.08}
    \resizebox{\columnwidth}{!}{
    \begin{tabular}{lcc}
        \hline
        \textbf{Method} &
        \textbf{Params (B)} &
        \textbf{LIBERO-Long} $\uparrow$ \\
        \hline
        OpenVLA               ~\cite{kim2024openvla}      & 7.0    & 53.7 \\
        UniAct                ~\cite{uniact2025}         & 0.5    & 46.5 \\
        UniAct + ACT Head     ~\cite{uniact2025,act2023} & $\sim$0.5 & 70.0 \\
        RoboMonkey            ~\cite{robomonkey2025}     & $\sim$14     & 56.5 \\
        PDF                   ~\cite{zang2026pdf}        & $7.0^{\dagger}$ & 59.0 \\
        Dita (OXE)            ~\cite{dita2025}           & 0.334  & 63.8 \\
        Dita (DROID)          ~\cite{dita2025}           & 0.334  & 77.9 \\
        ThinkAct              ~\cite{thinkact2025}       & 7.0    & 70.9 \\
        MolmoAct              ~\cite{molmoact2025}       & 7.0    & 77.2 \\
        Fast-ThinkAct         ~\cite{fastthinkact2026}   & 3.0    & \textbf{79.4} \\
        \hline
        \textbf{PLaCA (Ours)} & \textbf{0.395}           & \textbf{79.0} \\
        \hline
    \end{tabular}
    }
\end{table}

As shown in Table~\ref{tab:libero_long_comparison},
PLaCA achieves a success rate of 79.0\% on LIBERO-Long,
outperforming the reported results of several representative
methods, including Dita, ThinkAct, MolmoAct,and PDF, while
achieving performance comparable to Fast-ThinkAct (79.4\%).
Notably, PLaCA contains only 394.83M parameters, of which
11.66M (2.95\%) are trainable during the second training
stage. By directly predicting spatially structured future
latent change and using it for action generation, PLaCA
incorporates future information without generating complete
future images or intermediate motion sequences. This design
provides a practical way to incorporate future modeling
within a relatively small model, achieving competitive
closed-loop performance with a limited number of trainable
parameters.

Compared with the Action-only policy built on the same
base architecture, PLaCA improves the average success
rate from 74.0\% to 79.0\%, a gain of 5.0 percentage
points. The contributions of individual
components are further examined through ablation studies in Sec.~\ref{sec:ablation}.

\begin{figure}[t]
    \centering
    \includegraphics[width=\linewidth]{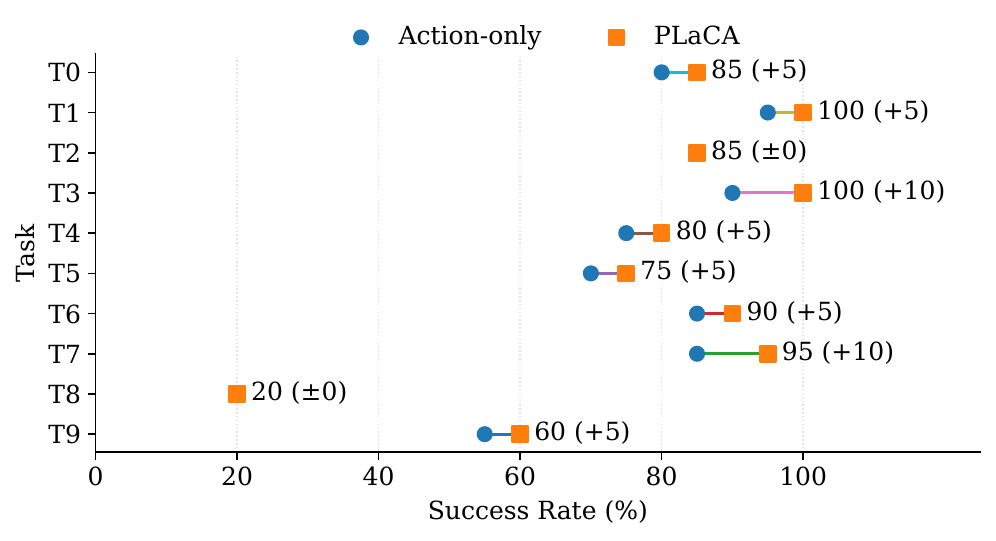}
    
    \caption{
    \textbf{Per-task success rates on LIBERO-Long.}
    Comparison between the Action-only baseline and
    PLaCA across all 10 tasks, with 20 closed-loop
    rollouts per task.
    }
    \label{fig:libero_per_task}
\end{figure}

\noindent\textbf{Per-Task Performance.}
To examine the performance across different manipulation
tasks, Fig.~\ref{fig:libero_per_task} compares the
per-task success rates of the Action-only policy and
PLaCA on all 10 LIBERO-Long tasks.

PLaCA improves the success rate on 8 of the 10 tasks
and maintains the same performance on the remaining
two tasks. In particular, PLaCA achieves a 100\%
success rate on Tasks 1 and 3. However, Task 8,
which requires placing two moka pots on the stove,
remains challenging, with a success rate of 20\%.

\noindent\textbf{Results on CALVIN.}
To further assess PLaCA on sequential robotic manipulation,
we evaluate its performance on CALVIN under the
ABC$\rightarrow$D setting. As shown in
Table~\ref{tab:calvin}, PLaCA achieves an ASL of 3.564,
outperforming the Action-only baseline (2.784) by 0.780.
This result demonstrates that PLaCA also improves
sequential task execution in an unseen environment.

\begin{table}[t]
    \centering
    \caption{
        \textbf{Results on CALVIN ABC$\rightarrow$D.}
        ASL denotes the average number of consecutively
        completed subtasks per five-task sequence.
    }
    \label{tab:calvin}
    \setlength{\tabcolsep}{12pt}
    \renewcommand{\arraystretch}{1.08}
    \begin{tabular}{lc}
        \hline
        \textbf{Method} & \textbf{ASL} $\uparrow$ \\
        \hline
        Action-only & 2.784 \\
        \textbf{PLaCA (Ours)} & \textbf{3.564} \\
        \hline
    \end{tabular}
\end{table}

\subsection{Ablation Studies}
\label{sec:ablation}

We conduct ablation studies on LIBERO-Long to examine
three key designs in PLaCA: patch-level future latent
change modeling, patch-level change-aware supervision,
and the future-to-action interface.
All variants are evaluated on the same 10 tasks with
20 closed-loop rollouts per task.
The results are summarized in
Table~\ref{tab:placa_ablation}.

\begin{table}[t]
    \centering
    \caption{
    \textbf{Ablation studies of PLaCA on LIBERO-Long.}
    Patch Grid denotes the spatial resolution of the
    future latent representation.
    Uniform and Change-aware indicate the patch-wise
    supervision weighting scheme.
    F2A denotes the future-to-action interface.
    }
    \label{tab:placa_ablation}
    \setlength{\tabcolsep}{4pt}
    \renewcommand{\arraystretch}{1.12}
    \resizebox{\columnwidth}{!}{
    \begin{tabular}{lcccc}
        \hline
        \textbf{Variant} &
        \textbf{Patch Grid} &
        \textbf{Supervision} &
        \textbf{F2A} &
        \textbf{Success} $\uparrow$ \\
        \hline
        Action-only
            & -- & -- & -- & 74.0 \\
        Action-only (+40 epochs)
            & -- & -- & -- & 72.5 \\
        PLaCA ($7\times7$)
            & $7\times7$ & Change-aware
            & \checkmark & 77.5 \\
        Patch-change $\rightarrow$ Action
            & $14\times14$ & Uniform
            & \checkmark & 76.0 \\
        Future-change auxiliary
            & $14\times14$ & Change-aware
            & -- & 76.5 \\
        Absolute future latent
            & $14\times14$ & Change-aware
            & \checkmark & 74.5 \\
        \hline
        \textbf{PLaCA (Ours)}
            & $\mathbf{14\times14}$
            & \textbf{Change-aware}
            & \checkmark
            & \textbf{79.0} \\
        \hline
    \end{tabular}
    }
\end{table}

\noindent\textbf{Effect of Additional Training.}
To examine whether performance gains arise from
additional training, we continue training the Action-only
policy for 40 epochs from the same checkpoint, matching
the second-stage training duration of PLaCA.
The extended Action-only baseline achieves 72.5\%,
compared with 74.0\% for the original baseline and
79.0\% for PLaCA.
This indicates that the improvement is attributable
to future latent change modeling rather than
additional training alone.

\noindent\textbf{Patch-Level Future Latent Change Modeling.}
We examine this design from two perspectives:
the prediction target and spatial structure.
First, we replace future latent change with absolute
future latent as the prediction target, while retaining
the same future-to-action interface and training
configuration. Directly predicting absolute future
latent achieves 74.5\%, 4.5 percentage points below
PLaCA (79.0\%), demonstrating the benefit of explicitly
modeling future latent change.

Second, we investigate spatial structure by varying
the patch grid resolution, preserving patch-wise
representations and supervision rather than removing
them altogether. Reducing the grid from $14\times14$
to $7\times7$ decreases the success rate from
79.0\% to 77.5\%, suggesting that retaining richer
local spatial change information benefits action
generation.

\noindent\textbf{Effect of Patch-Level Change-Aware Supervision.}
Replacing change-aware supervision with uniform
patch weighting reduces the success rate from
79.0\% to 76.0\%.
This demonstrates the benefit of adjusting supervision
weights according to local future latent change
magnitudes, enabling more effective learning of
future environmental changes.

\noindent\textbf{Effect of the Future-to-Action Interface.}
Using future latent change prediction only as an
auxiliary training objective achieves 76.5\%.
Explicitly incorporating the predicted future latent
change into action generation through the
future-to-action interface improves the success
rate to 79.0\%, confirming the effectiveness of
directly coupling future modeling with action generation.

\section{Conclusion}
\label{sec:conclusion}

In this paper, we propose PLaCA, a world-to-action framework built around future latent change, aiming to make future modeling more directly contribute to robot action generation. By preserving fine-grained spatial structure, employing change-aware supervision, and explicitly incorporating predicted future latent change into action generation, PLaCA establishes a direct connection between future modeling and action generation. Experiments on LIBERO-Long and CALVIN demonstrate improved downstream manipulation performance, while ablation studies further validate the effectiveness of the proposed designs. Notably, PLaCA achieves performance improvements by updating only a small fraction of model parameters during the second training stage, suggesting that future modeling can effectively enhance action generation with limited parameter updates. Overall, our results highlight spatially structured future latent change as an effective representation for connecting future modeling with robot action generation.

{
    \small
    \bibliographystyle{ieeenat_fullname}
    \bibliography{main}
}


\end{document}